\pdfoutput=1
\documentclass[letterpaper, 10 pt, conference]{ieeeconf}  

\IEEEoverridecommandlockouts                              
\usepackage{amsmath} 
\usepackage{amssymb}  

\usepackage{graphicx}
\usepackage{bm}
\usepackage{pgfplots}
\usepackage{float}
\usepackage{subfig}
\usepackage[noadjust]{cite}
\usepackage{dblfloatfix}
\usepackage{array}
\usepackage{wrapfig}

\newcommand{\thickhline}{%
    \noalign {\ifnum 0=`}\fi \hrule height 1pt
    \futurelet \reserved@a \@xhline
}
\newcolumntype{"}{@{\hskip\tabcolsep\vrule width 1pt\hskip\tabcolsep}}
\makeatother

\title{\LARGE \bf
Skill Acquisition via Automated Multi-Coordinate Cost Balancing
}

\author{Harish Ravichandar$^1$$^{\dagger}$, S. Reza Ahmadzadeh$^2$$^\dagger$, M. Asif Rana$^1$, and Sonia Chernova$^1$
\thanks{$^\dagger$ indicates equal contribution}%
\thanks{$^1$ Georgia Inst. of Technology, Atlanta, GA. Email: {\tt\small \{harish. ravichandar,asif.rana,chernova\}@gatech.edu}}%
\thanks{$^2$ University of Massachusetts Lowell, Lowell, MA. Email: {\tt\small reza\_ahmadzadeh@uml.edu}}%
}

\begin{document}

\maketitle
\thispagestyle{empty}
\pagestyle{empty}

\begin{abstract}
We propose a learning framework, named Multi-Coordinate Cost Balancing (MCCB), to address the problem of acquiring point-to-point movement skills from demonstrations. MCCB encodes demonstrations simultaneously in multiple differential coordinates that specify local geometric properties. MCCB generates reproductions by solving a convex optimization problem with a multi-coordinate cost function and linear constraints on the reproductions, such as initial, target, and via points. Further, since the relative importance of each coordinate system in the cost function might be unknown for a given skill, MCCB learns optimal weighting factors that balance the cost function. We demonstrate the effectiveness of MCCB via detailed experiments conducted on one handwriting dataset and three complex skill datasets.
\end{abstract}

\section{Introduction}
\label{sec:intro}

The next generation of robots, that can operate in and adapt to unstructured and dynamic environments, must possess a diverse set of skills. However, it is implausible to pre-program robots with a library of all required skills. Learning from Demonstration (LfD) \cite{argall2009survey,billard2008robot} is a paradigm that aims to equip robots with the ability to learn efficiently from demonstrations provided by humans. Existing work in trajectory-based LfD has contributed a wide range of mathematical representations that encode skills from human demonstrations and then reproduce the learned skills at runtime. Proposed representations include Spring-damper systems with forcing functions \cite{pastor2009learning}, Gaussian Mixture Models (GMMs) \cite{calinon2007learning,khansari2011learning,ravichandar2018learning}, Neural Networks (NNs) \cite{neumann2013neural,levine2014learning}, Gaussian Processes (GPs) \cite{schneider2010robot,rana2017towards,umlauft2017learning}, and geometric objects \cite{ahmadzadeh2017generalized}, among others. Each of these representations is used to encode the demonstrations in a predefined space or coordinate system (e.g., Cartesian coordinates). In other words, a single best coordinate system for any given skill is assumed to both exist and be known. However, as we show in this work, the assumption that a single best coordinate system exists for each task does not hold. Further, encoding in only a single coordinate system prohibits the model from capturing some of the geometric features that underlie a demonstrated skill.

\begin{figure}
  \centering
  \includegraphics[trim={6cm 1.1cm 6cm 1.2cm}, clip, width=\columnwidth]{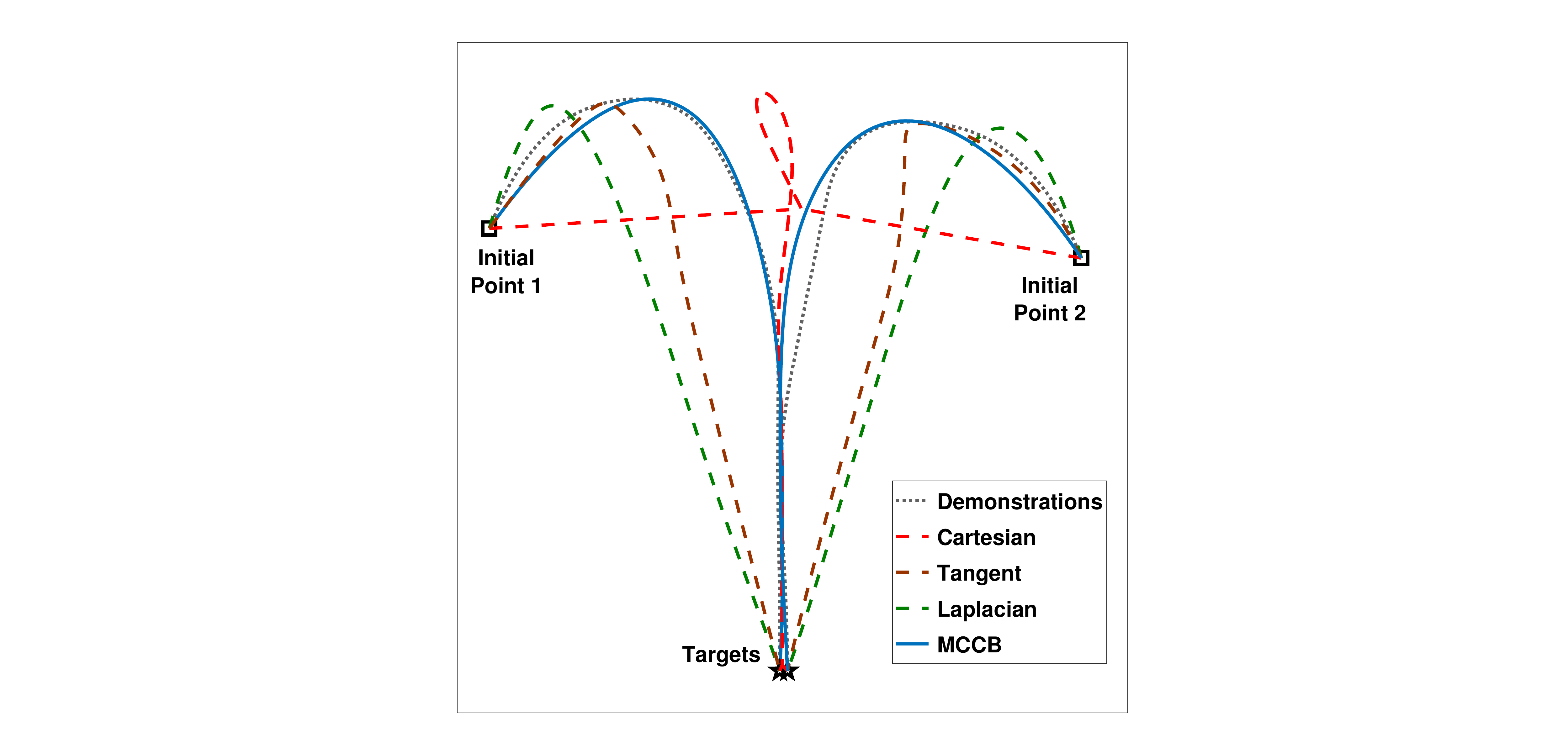}
  \caption{\small{A comparison of reproductions generated by considering different coordinates, illustrating the need for cost balancing.}}
  \label{fig:converging_traj}
\end{figure}

In this work, we contribute a learning framework that encodes demonstrations simultaneously in multiple coordinates, and balances the relative influences of the learned models in generating reproductions. The proposed framework, named Multi-Coordinate Cost Balancing (MCCB), encodes demonstrations in three differential coordinates: Cartesian, tangent, and Laplacian (Section \ref{subsec:coordinate_transformations}). Simultaneously learning in these three coordinates allows our method to capture all of the underlying geometric properties that are central to a given skill. MCCB encodes the joint density of the time index and the demonstrations in each differential coordinate frame using a separate statistical model. Thus, given any time instant, we are able to readily obtain the conditional mean and covariance in each coordinate system (Section \ref{subsec:GMMs}). MCCB generates reproductions by solving an optimization problem with a blended cost function that consists of one term per coordinate. Each term penalizes deviations from the norm, weighted by the inverse of the expected variance in the corresponding coordinate system (Section \ref{subsec:optimization}). Further, we subject the optimization problem to linear constraints on the reproductions, such as initial, target, and via point constraints. Our constrained optimization problem is convex with respect to the reproduction and hence can be solved efficiently.

\begin{figure}
  \centering
  \includegraphics[trim={0.5cm 0.5cm 0.5cm 0.5cm}, clip, width=\columnwidth]{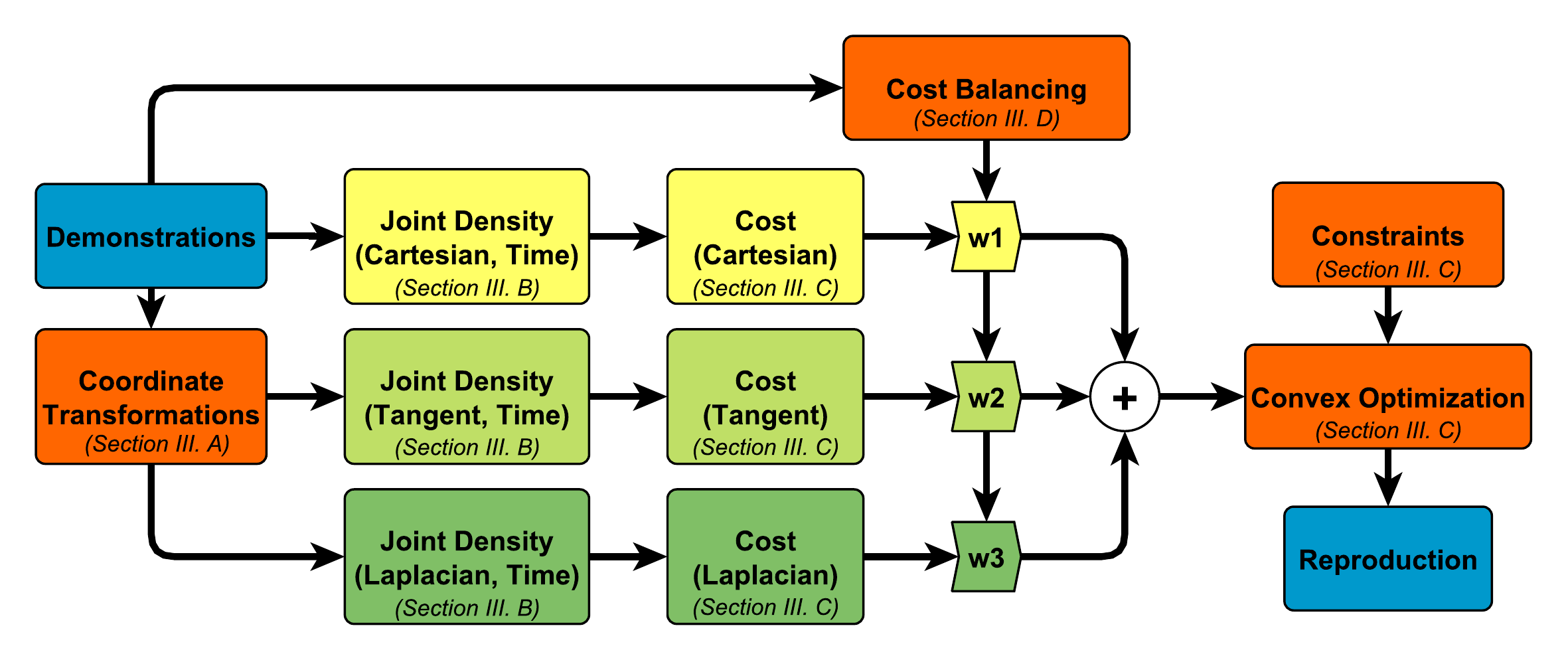}
  \caption{\small{A flow diagram illustrating MCCB.}}
  \label{fig:graph}
\end{figure}

A major hurdle in learning a wide variety of skills, without significant parameter tweaking, is that the relative importance of each differential coordinate (or the geometric feature) in encoding a given task is unknown ahead of time. For instance, consider the problem of encoding the demonstrations illustrated in Fig. \ref{fig:converging_traj}. Using any one coordinate system in isolation, even when the most suitable one is known, does not yield good reproductions (the red, brown, and green dashed lines). To alleviate this problem, MCCB preferentially weights the costs defined in each coordinate (Fig.~\ref{fig:graph}). Importantly, MCCB learns the optimal weights directly from the demonstrations without making task-dependent assumptions. To this end, MCCB solves a meta optimization problem that aims to minimize reproduction errors (Section \ref{subsec:cost_balancing}). As shown by the solid blue lines in Fig. \ref{fig:converging_traj}, a cost function that optimally balances the costs in each coordinate yields better reproductions than any single-coordinate method.

In summary, we contribute a unified task-independent learning framework that (1) encodes demonstrations simultaneously in multiple differential coordinates, (2) defines a blended cost function that incentivizes conformance to the norm in each coordinate system while considering expected variance, and (3) learns optimal weights directly from the demonstrations to balance the relative influence of each differential coordinate in generating reproductions. Further, MCCB is compatible with and complementary to several existing LfD methods that utilize different statistical representations and coordinate systems \cite{calinon2014task,paraschos2013probabilistic,ahmadzadeh2017generalized,umlauft2017learning,rana2017towards,osa2017guiding,nierhoff2016spatial}.

\section{Related Work}
\label{sec:relatedwork}

Learning from demonstration has attracted a lot of attention from researchers in the past few decades. While several categories of LfD methods exist \cite{argall2009survey}, our work falls under the category of trajectory-based LfD. In this category, demonstrations take the form of trajectories and the methods aim to synthesize trajectories that accurately reproduce the demonstrations.

Dynamical systems-based trajectory learning methods, such as \cite{khansari2011learning,ravichandar2018learning,neumann2013neural}, encode demonstrations using statistical dynamical systems and generate reproductions by forward propagating the dynamics. While such deterministic methods exhibit several advantages, such as convergence guarantees and robustness to perturbations, they are restricted to learning in a single coordinate system and ignore inherent uncertainties in the demonstrations. They incentivize conformance to the norm even when demonstrations exhibit high variance.

Trajectory optimization methods, such as \cite{ratliff2009chomp} and \cite{schulman2014motion}, focus on geometric features by minimizing costs specified using predefined norms. An optimization framework proposed in \cite{dragan2015movement} attempts to adapt multiple demonstrations to new initial and target locations by minimizing the distance between the demonstrations and the reproduction according to a learned Hilbert space norm. Indeed, learning an appropriate Hilbert space norm is related to finding an appropriate coordinate system based on the demonstrations. However, similar to the dynamical systems-based methods, the methods in \cite{ratliff2009chomp,schulman2014motion,dragan2015movement} are restricted to a single predefined or learned coordinate system and do not explicitly model and utilize the inherent time-dependent variations in the demonstrations.

Probabilistic trajectory-learning methods, such as \cite{rana2017towards,umlauft2017learning} and \cite{paraschos2013probabilistic}, on the other hand, capture and utilize the variation observed in the demonstrations. However, these methods are also restricted to encoding demonstrations in a single predefined coordinate system that is assumed to be known.

Our design of the costs in each differential coordinate is inspired by the minimal intervention principle \cite{calinon2014task} that takes variance into account. While the approach in \cite{calinon2014task} does encode demonstrations in different frames of references, all the frames are restricted to Cartesian coordinates or orientation space. Furthermore, all the relevant frames for a given task are also expected to be provided by the user.

The motion planning framework in \cite{osa2017guiding}, complementary to our approach, utilizes a blended cost function, the construction of which is guided by probability distributions learned from the demonstrations. This framework incentivizes factors such as smoothness, manipulability, and obstacle avoidance, but is restricted to the Cartesian coordinate system. MCCB, on the other hand, encodes demonstrations in multiple differential coordinates and learns to optimally balance their relative influences, but does not consider factors such as manipulability and obstacle avoidance.

Differential coordinates have been extensively used in the computer graphics community \cite{lipman2004differential,levy2006laplace}. Prior work in trajectory learning that incorporates differential coordinates includes the Laplacian trajectory editing (LTE) algorithm \cite{nierhoff2016spatial}. Using Laplacian coordinates, the  LTE algorithm adapts a single demonstration to new initial, target, and via points while preserving the shape. However, the LTE algorithm does not reason about the relative importances of multiple coordinates.

\section{Methodology}\label{sec:method}

The section describes the technical details of MCCB and its work flow as illustrated in Fig. \ref{fig:graph}.
\subsection{Differential Coordinate Transformations} \label{subsec:coordinate_transformations}
In this section, we define the differential coordinates and their corresponding transformations used in MCCB.

\textit{Cartesian:} Let a discrete finite-length trajectory in $n$-dimensional \textit{Cartesian coordinates} be denoted by $\bm{X} = [x(1) \ x(2) \cdots x(T)]^\top \in \mathbb{R}^{T \times n}$ and let $x(t) \in \mathbb{R}^n$ denote a discrete sample at time index $t$. This trajectory can be represented using a graph $\mathcal{G}=(\mathcal{V},\mathcal{E})$ where $\mathcal{V}$ is the set of vertices representing the samples in the trajectory and $\mathcal{E}$ is the set of edges that represent the connections between the samples in the trajectory. The neighborhood $\mathcal{N}_t$ of each vertex $\mathcal{V}_t$ is defined by the set of adjacent vertices $\mathcal{V}_t'$. In the case of discrete-time trajectories, the edges between any given vertex and its two neighbors are assumed to carry unit weights, while all other edges carry zero weights.

\textit{Laplacian:} It is known that the discrete Laplace-Beltrami operator for the trajectory $\bm{X}$ provides the \textit{Laplacian coordinate} $\delta(t)$ as
${
\delta(t) \triangleq \sum_{t' \in \mathcal{N}_t} \frac{1}{\sum_{t' \in \mathcal{N}_t} 1} (x(t) - x(t'))
}$ \cite{lipman2004differential}.
Note that the above relationship can be written as a linear differential operator in matrix form
\begin{equation}\label{eq:Delta}
\bm{\Delta} = \bm{L}\bm{X}
\end{equation}
where $\bm{\Delta} = [\delta(1) \  \delta(2) \cdots \delta(T)]^\top \in \mathbb{R}^{T \times n}$ is the trajectory in the Laplacian coordinates, and $\bm{L} \in \mathbb{R}^{T \times T}$, called the graph Laplacian, is given by
\begin{equation}
\bm{L} = \left[\begin{smallmatrix}
    		  1     & -1     &  0     & \dots  & \dots  &  0     \\
   			-0.5    &  1     & -0.5   &  0     & \dots  &  0     \\
    		  0     & -0.5   &  1     & -0.5   & \dots  &  0     \\
    	   \vdots   &        & \ddots & \ddots & \ddots & \vdots \\
              0     & \dots  &  0     & -0.5   &  1     & -0.5   \\
              0     & \dots  & \dots  &  0     & -1     &  1
		\end{smallmatrix}\right]
\end{equation}
As pointed out in \cite{nierhoff2016spatial}, the Laplacian coordinates have meaningful geometric interpretations. Specifically, the Laplacian coordinates can be seen as the discrete approximations of the derivative of the unit tangent vectors of an arc-length parametrized continuous trajectory. In other words, the Laplacian coordinates measure the deviation of each sample from the centroid of its neighbors.

\textit{Tangent:} While the Laplacian coordinates are discrete approximations of second order differential transformations, a discrete approximation of the first differential transformation is possible. Consider such a first order transformation using first order finite differences defined as
${
\gamma(t) \triangleq (x(t+1) - x(t))
}$,
where $\gamma(t)$ is called the \textit{tangent coordinate}. The matrix form of the above relationship results in a linear differential operator given by
\begin{equation} \label{eq:Gamma}
\bm{\Gamma} = \bm{G}\bm{X}
\end{equation}
where $\bm{\Gamma} = [\gamma(1) \  \gamma(2) \cdots \gamma(T)]^\top \in \mathbb{R}^{T \times n}$ is the trajectory in the tangent coordinates and $\bm{G} \in \mathbb{R}^{T \times T}$, called the graph incidence matrix, is given by
\begin{equation}
\bm{G} = \left[\begin{smallmatrix}
    		  -1    &  1     &  0     & \dots  & \dots  &  0     \\
   			   0    & -1     &  1     &  0     & \dots  &  0     \\
    		   0    &  0     & -1     &  1     & \dots  &  0     \\
    	   \vdots   &        & \ddots & \ddots & \ddots & \vdots \\
               0    & \dots  &  0     &  0   & -1       &  1     \\
               0    & \dots  & \dots  &  0     &  0     & -1
		\end{smallmatrix}\right]
\end{equation}
Similar to the Laplacian coordinates, the tangent coordinates have geometric interpretations. Specifically, the tangent coordinates can be seen as discrete approximations of the un-normalized tangent vectors of an arc-length parametrized continuous trajectory, i.e., the tangent coordinates measure the local direction of motion at each sample of the trajectory.

In our work, we assume that a set of $N$ demonstrations in the Cartesian coordinates are available. Let the $j$th demonstration be denoted by $\bm{X}_d^j = [x_d^j(1) \ x_d^j(2) \cdots x_d^j(T)]^\top \in \mathbb{R}^{T \times n}$. Note that if the raw demonstrations are of varying duration in time, we perform time alignment using dynamic time warping. MCCB transforms each obtained demonstration $\bm{X}_d^j$ into a trajectory in the tangent coordinates (denoted by $\bm{\Gamma}_d^j$) and a trajectory in Laplacian coordinates (denoted by $\bm{\Delta}_d^j$) using (\ref{eq:Delta}) and (\ref{eq:Gamma}), respectively.

\subsection{Encoding in Multiple Differential Coordinates}\label{subsec:GMMs}

This section defines the costs associated with each coordinate. With the demonstrations available in all three differential coordinates, we employ three independent Gaussian mixture models (GMMs)\footnote{MCCB does not rely on the use of GMMs and any statistical representation that can provide the conditional estimates will suffice.} to approximate the joint probability densities of time and the samples in each coordinate system.

The GMM associated with the Cartesian coordinates attempts to approximate the joint density of $t$ and $x$ using a finite number of Gaussian basis functions as follows
${
\mathcal{P}(t,x;\theta_{C}) = \sum_{k=1}^{K_C} \mathcal{P}(k) \mathcal{P}(t,x|k)
}$,
where $K_C$ is the number of Gaussian basis functions, $\mathcal{P}(k) = \pi_{C}^k$ is the prior associated with the $k$th basis function, $\theta_{C} = \{\mu_{C}^1 \cdots \mu_{C}^{K_{C}}, \Sigma_{C}^1 \cdots \Sigma_{C}^{K_{C}}, \pi_{C}^1 \cdots \pi_{C}^{K_{C}}\}$ is the set of parameters of the GMM, and $\mathcal{P}(t,x|k)$ is the conditional probability density given by
${
\mathcal{P}(t,x|k) \sim \mathcal{N}\left(
\begin{bmatrix} t \\ x \end{bmatrix};
\mu_{C}^k,\Sigma_{C}^k\right)
}$,
where $\mu_{C}^k = \begin{bmatrix} \mu_t^k \\ \mu_{x}^k \end{bmatrix}$ is the mean and $\Sigma_{C}^k = \begin{bmatrix} \Sigma_{t}^k & \Sigma_{t,x}^k \\ \Sigma_{x,t}^k & \Sigma_{x}^k \end{bmatrix}$ is the covariance matrix of the $k$th Gaussian basis function.

We learn the parameters $\theta_C$ of the model using the Expectation-Maximization algorithm based on the demonstrations $\{\bm{X}_d^j\}_{j=1}^{N}$. Given the learned model and a time instant, the expected value of the conditional density $\mathcal{P}(x|t)$ is given by Gaussian mixture regression (GMR) \cite{cohn1996active} as follows

\begin{equation}\label{eq:x_hat}
\hat{x}(t) = \mathbb{E}[x|t] = \sum_{k=1}^{K_C} h_C^k(t) (A_C^k t + b_C^k)
\end{equation}
where $h_C^k(t) = \frac{\mathcal{P}(k) \mathcal{P}(t|k)}{\sum_{i=1}^{K_C}\mathcal{P}(i) \mathcal{P}(t|i)}$, $A_C^k = \Sigma_{x,t}^k (\Sigma_{t}^k)^{-1}$, $b^k = \mu_{x}^k +  (t - \mu_t^k)$, and the conditional covariance is given by
\begin{equation}\label{eq:Sigma_x_hat}
\hat{\Sigma}_{x}(t) = Var[x|t] = \sum_{k=1}^{K_C} {h_C^k}^2 \ (\Sigma_{x}^k - \Sigma_{x,t}^k(\Sigma_{t}^k)^{-1}\Sigma_{t,x})
\end{equation}

Similar to the GMM learned in the Cartesian coordinates, we learn a second GMM in the tangent coordinates based on the demonstrations $\{\bm{\Gamma}_d^j\}_{j=1}^{N}$, and a third GMM in the Laplacian coordinates based on the demonstrations $\{\bm{\Delta}_d^j\}_{j=1}^{N}$. The expected values of the conditional densities $\mathcal{P}(\gamma|t)$ and $\mathcal{P}(\delta|t)$ are given by
\begin{align}
\hat{\gamma}(t) = & \mathbb{E}[\gamma|t] = \sum_{k=1}^{K_G} h_G^k(t) (A_G^k t + b_G^k) \label{eq:gamma_hat} \\
\hat{\delta}(t) = & \mathbb{E}[\delta|t] = \sum_{k=1}^{K_L} h_L^k(t) (A_L^k t + b_L^k) \label{eq:delta_hat}
\end{align}
and the corresponding conditional expectations are given by
\begin{align}
\hat{\Sigma}_{\gamma}(t) =& \mathrm{Var}[\gamma|t] = \sum_{k=1}^{K_G} ({h_G^k})^2 \ (\Sigma_{\gamma}^k -\Sigma_{\gamma,t}^k(\Sigma_{t}^k)^{-1}\Sigma_{t,\gamma}) \label{eq:Sigma_gamma_hat} \\
\hat{\Sigma}_{\delta}(t) =& \mathrm{Var}[\delta|t] = \sum_{k=1}^{K_L} ({h_L^k})^2 \ (\Sigma_{\delta}^k -\Sigma_{\delta,t}^k(\Sigma_{t}^k)^{-1}\Sigma_{t,\delta}) \label{eq:Sigma_delta_hat}
\end{align}
where the variables in (\ref{eq:gamma_hat})-(\ref{eq:Sigma_delta_hat}) with subscripts $G$ and $L$ correspond to the tangent and Laplacian coordinates, respectively, and are defined similarly to the ones in (\ref{eq:x_hat})-(\ref{eq:Sigma_x_hat}).

\subsection{Imitation via Optimization}\label{subsec:optimization}
In this section, we explain the design of our multi-coordinate cost function. MCCB generates reproductions by solving a constrained optimization problem given by
\begin{align}
\bm{X}_r = & \arg \min_{\bm{X}} \ w_C J_C(\bm{X}) + w_G J_G(\bm{X}) \nonumber \\
                                  & \qquad \qquad \qquad + w_L J_L(\bm{X}) \label{eq:X_opt} \\
 		   &\ \mathrm{s.t.} \qquad \ P_x \bm{X} = \bm{X}^* \label{eq:X_opt_P}
\end{align}
where $\bm{X}_r \in \mathbb{R}^{T \times n}$ is the reproduction, $w_C,\ w_G,\ w_L \in \mathbb{R}^+$ are positive weights; $J_C,\ J_G,\ J_L: \mathbb{R}^{T \times n} \rightarrow \mathbb{R}^+$ are cost functions in the Cartesian, tangent, and Laplacian coordinates, respectively; $P_x \in \mathbb{R}^{m \times T}$ and $\bm{X}^* \in \mathbb{R}^{m \times n}$ define $m \in \mathbb{Z}^+$ linear constraints on $\bm{X}_r$. In practice, $m<<n$ and we use the linear constraints to enforce constraints on initial, target, and via points.

We define the cost function in each coordinate system as follows
\begin{align}
J_C(\bm{X}) = & (\bm{X}(:) - \hat{\bm{X}}(:))^\top (\bm{\hat{\Sigma}_{\bm{X}}})^{-1}  (\bm{X}(:) - \hat{\bm{X}}(:)) \label{eq:Cartesian_cost} \\
J_G(\bm{X}) = & (\bm{\Gamma}(:) - \hat{\bm{\Gamma}}(:))^\top (\bm{\hat{\Sigma}_{\Gamma}})^{-1}  (\bm{\Gamma}(:) - \hat{\bm{\Gamma}}(:)) \label{eq:tangent_cost} \\
J_L(\bm{X}) = & (\bm{\Delta}(:) - \hat{\bm{\Delta}}(:))^\top (\bm{\hat{\Sigma}_{\Delta}})^{-1}  (\bm{\Delta}(:) - \hat{\bm{\Delta}}(:))  \label{eq:Laplacian_cost}
\end{align}
where $\bm{\hat{\Sigma}_{\bm{X}}}, \bm{\hat{\Sigma}_{\Gamma}}, \bm{\hat{\Sigma}_{\Delta} \in \mathbb{R}^{nT \times nT}}$ denote the block diagonal matrices formed with the conditional covariances $\hat{\Sigma}_x (t), \hat{\Sigma}_\gamma (t),$ and $\hat{\Sigma}_\delta (t)$, respectively, for all values of $t$. Further, the notation $(:)$ denotes vectorization - for instance, $\bm{X}(:), \hat{\bm{X}}(:) \in \mathbb{R}^{nT}$ denote the vectorized trajectories formed by vertically stacking $x(t)$ and $\hat{x}(t)$ for all values of $t$, respectively. Note that we construct the trajectories $\Gamma$ and $\Delta$ in (\ref{eq:tangent_cost}) and (\ref{eq:Laplacian_cost}) from $\bm{X}$ via the linear operators defined in (\ref{eq:Gamma}) and (\ref{eq:Delta}), respectively. MCCB penalizes deviations from the conditional mean in each coordinate system. However, deviations are penalized less (more) severely if high (low) variance is observed in the demonstrations at any given time.

\subsection{Automated Cost Balancing}\label{subsec:cost_balancing}
In order to obtain reproductions that successfully imitate demonstrations of a wide variety of skills, the weights $w_C,\ w_G,$ and $w_L$ have to be chosen with care. Indeed, they preferentially weight the costs defined in each differential coordinate and thereby manipulate the relative incentive for successful imitation in each coordinate system.

We learn these weights directly from the available demonstrations. Note that, for known weights, the constrained optimization problem in (\ref{eq:X_opt}) is convex in $\bm{X}$. We estimate the weights in the following form
\begin{equation}
\hat{w}_C = \frac{\alpha_C}{\beta_C}; \quad \hat{w}_G = \frac{\alpha_G}{\beta_G}; \quad \hat{w}_L = \frac{\alpha_L}{\beta_L}
\end{equation}
where $\beta_C,\ \beta_G,\ \beta_L \in ( 0, 1 ]$, such that $\sum_i \beta_i = 1$, are positive scaling factors used to correct for inherent differences in the magnitudes of the costs, and $\alpha_C,\ \alpha_G,\ \alpha_L \in [ 0, 1 ]$, such that $\sum_i \alpha_i = 1$, are positive weights used to preferentially weight the cost defined in each coordinate system. MCCB estimates the scaling factors $\beta_i$'s as follows
\begin{equation}
\beta_i = \frac{\sum_{j=1}^{N} J_i(\bm{X}_d^j)}{\sum_l \sum_{j=1}^{N} J_l(\bm{X}_d^j)}, \quad \forall i,l = \{C,G,L\}
\end{equation}

With the scaling factors compensating the inherent scale difference in the costs, we compute the preferential weighting factors $\alpha_i$'s that minimize reproduction error. To this end, we formulate the following meta optimization problem
\begin{align}
\{\alpha_C, \alpha_G, \alpha_L\} = & \arg \min_{\alpha_C, \alpha_G, \alpha_L} \sum_{j=1}^N \mathrm{SSE}(\bm{X}_r^j,\bm{X}_d^j) \\
                                   & \mathrm{s.t.} \sum_i \alpha_i = 1,\ \forall i = \{C,G,L\}
\end{align}
where $\mathrm{SSE}(\cdot)$ denotes the sum of squared errors computed over time, and $\bm{X}_r^j$ is the solution to the following optimization problem
\begin{align}
\bm{X}_r^j = & \arg \min_{\bm{X}} \ \left(\frac{\alpha_C}{\beta_C}\right) J_C(\bm{X}) + \left(\frac{\alpha_G}{\beta_G}\right) J_G(\bm{X})  \nonumber \\
                                  & \qquad \qquad \qquad + \left(\frac{\alpha_L}{\beta_L}\right) J_L(\bm{X}) \label{eq:X_j_opt}  \\
 		   &\ \mathrm{s.t.} \qquad \ P_x \bm{X} = \bm{X}_j^* \label{eq:X_j_opt_P}
\end{align}
where $P_x \bm{X} = \bm{X}_j^*$ denotes specific linear constraints pertaining to the demonstration $\bm{X}_d^j$, such as initial, target, and via points. Solving the above meta-optimization problem results in the preferential weights $\alpha_i$'s that minimize reproduction errors of the solutions generated by the original constrained optimization problem in (\ref{eq:X_opt})-(\ref{eq:X_opt_P}).

\section{Experimental Evaluation}\label{sec:experiments}

This section describes the design and discusses the results of four experiments conducted to evaluate MCCB. In each experiment, we compared the performances of the following approaches:
\begin{enumerate}
\item\textit{Cartesian-coordinates}: $w_C = 1$, $w_G=0$, $w_L=0$
\item \textit{Tangent-coordinates}: $w_C = 0$, $w_G=1$, $w_L=0$
\item \textit{Laplacian-coordinates}: $w_C = 0$, $w_G=0$, $w_L=1$
\item \textit{Uniform weighting}: $w_C = 1/3$, $w_G=1/3$, $w_L=1/3$
\item \textit{MCCB}: $w_C = \hat{w_C}$, $w_G=\hat{w_G}$, $w_L=\hat{w_L}$
\end{enumerate}

\begin{figure}
  \centering
  \includegraphics[trim={17cm 6.8cm 17cm 3.3cm}, clip, width=\columnwidth]{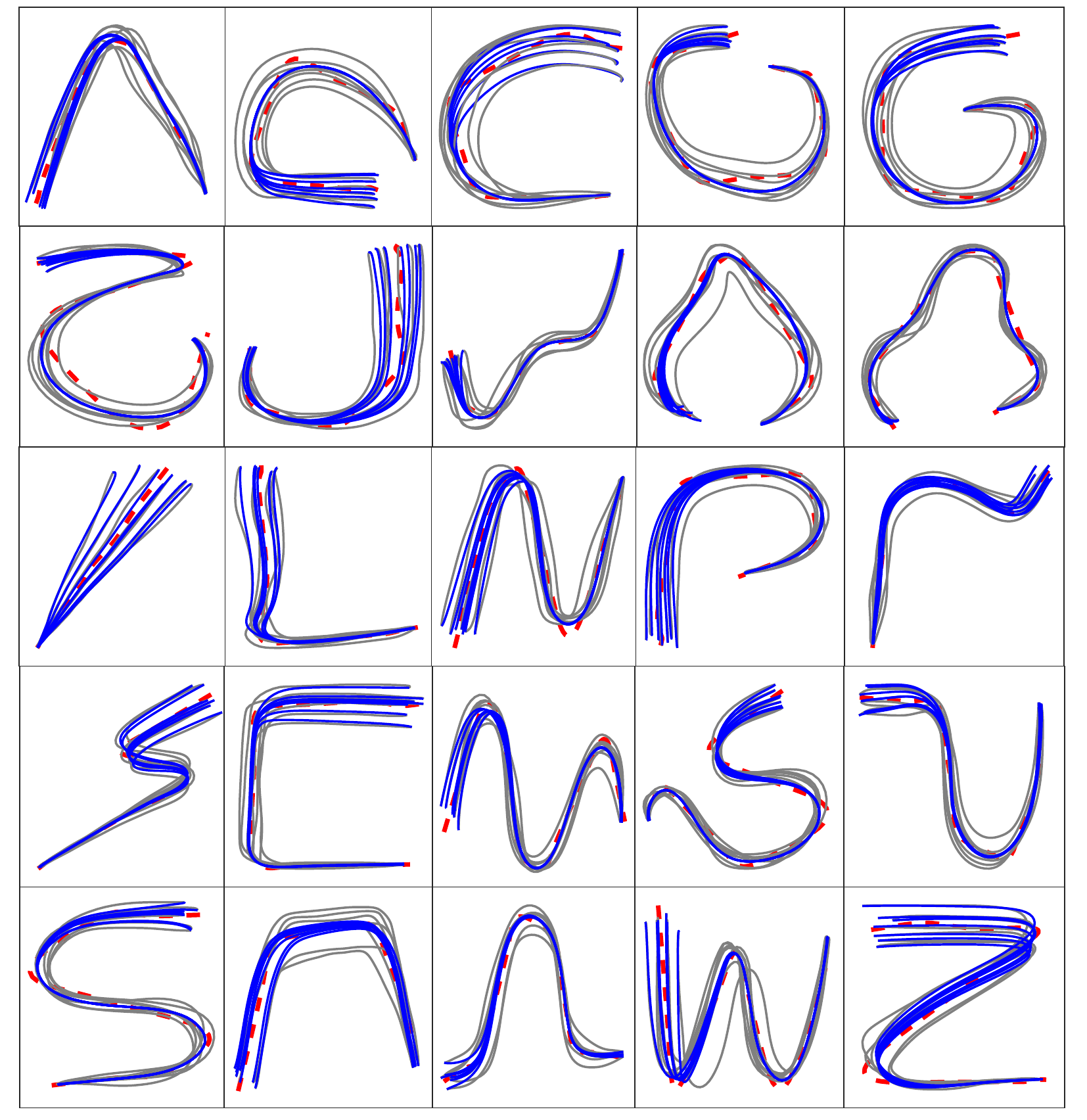}
  \caption{\small{Qualitative performance of MCCB on the LASA handwriting dataset. Demonstration (gray), reproductions (blue), and expected mean position (dashed red) are shown.}}
  \label{fig:LASA_qualitative}
\end{figure}

We measured the performance of each approach by the following geometric and kinematic metrics: \textit{Swept Error Area (SEA)} \cite{khansari2014learning}, \textit{Sum of Squared Errors (SSE)}, \textit{Dynamic Time Warping Distance (DTWD)}, and \textit{Frechet Distance (FD)} \cite{frechet1906quelques}. These metrics allow us to evaluate different aspects of each method's performance. The SEA and SSE metrics penalize both spatial and temporal misalignment, and thus evaluate kinematic performance. On the other hand, the DTWD and FD metrics penalize spatial misalignment while disregarding time misalignment, and thus evaluate geometric performance. Further, the SEA, SSE, and DTWD metrics evaluate aggregate performance by summing over or averaging across all the samples of each reproduction. The FD metric, on the other hand, computes the shortest possible cord length required to connect the demonstration and the reproduction in space while allowing time re-parametrization of either trajectory, and thus measures maximal deviation in space. Note that the SEA metric is restricted to 2-dimensional data, so we only report it for one of our experiments.

In all the experiments, we used the position constraints in (\ref{eq:X_opt_P}) to enforce both initial and end point constraints uniformly across all the methods being compared. Further, we uniformly set the number of Gaussian basis functions to five across all the coordinates and all the experiments.

\subsection{Handwriting Skill}\label{subsec:LASA}
This experiment evaluates MCCB on the publicly available LASA human handwriting library \cite{khansari2011learning}, that consists of handwriting motions collected from pen input using a Tablet PC. The library contains a total of 25 handwriting motions, each with 7 demonstrations. 

\begin{figure}
  \centering
  \includegraphics[trim={6.5cm 4cm 9cm 0cm}, clip, width=\columnwidth]{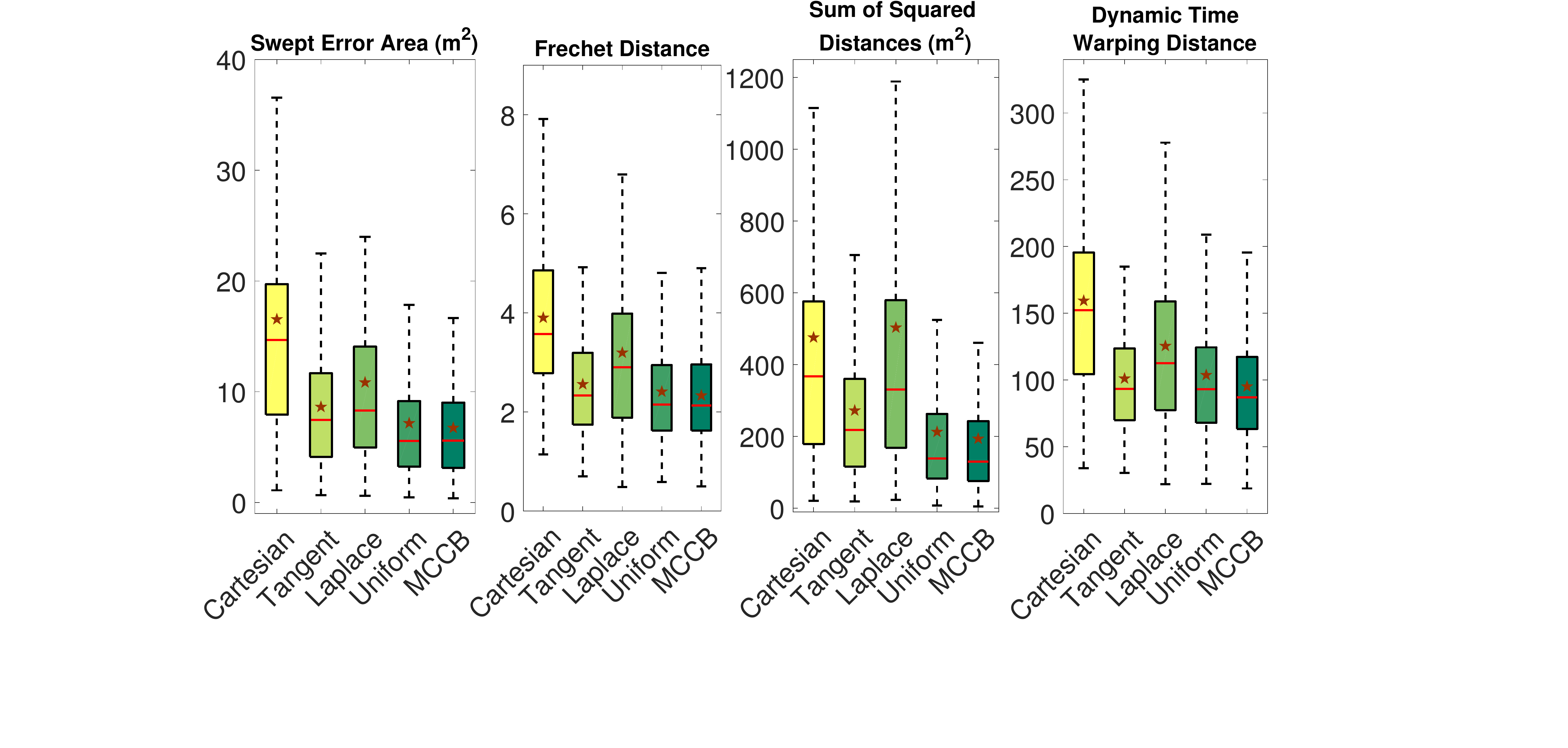}
  \caption{\small{Box plots, with mean (brown star) and median (red line), illustrate the performance of each approach on the handwriting task.}}
  \label{fig:LASA_quantitative}
\end{figure}

\begin{figure}
  \centering
  \includegraphics[trim={1cm 1cm 1cm 2cm}, clip, width=\columnwidth]{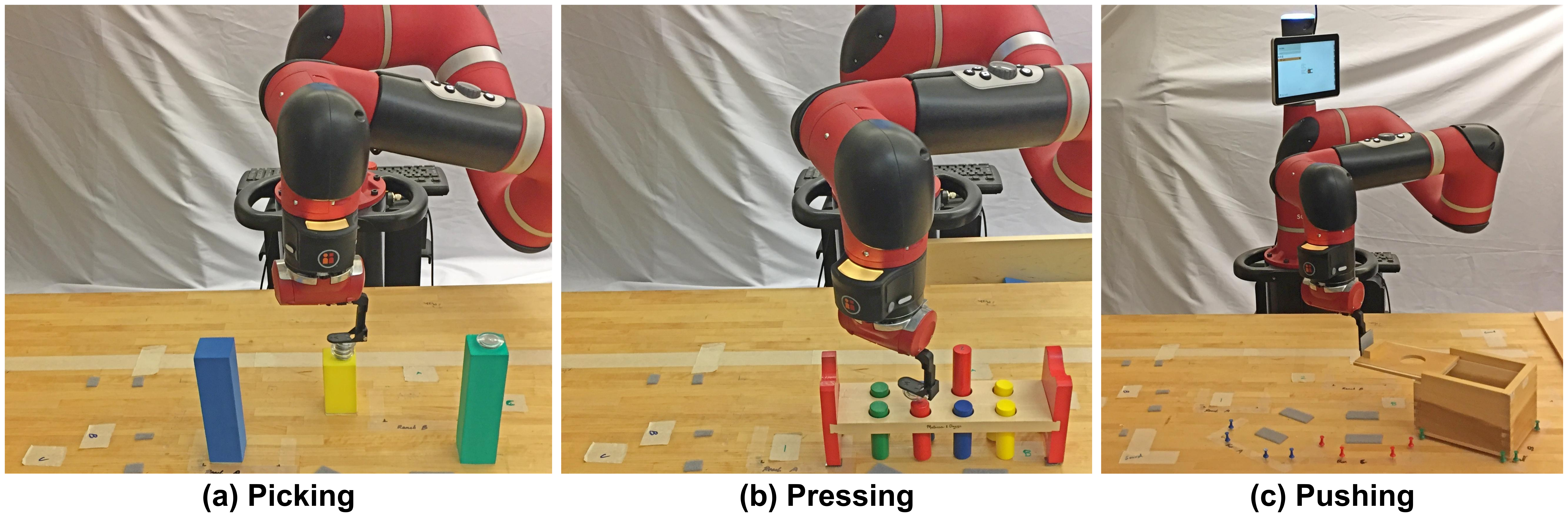}
  \caption{\small{Snapshots illustrating the experimental setup for the picking (left), pressing (center), and pushing (right) skills.}}
  \label{fig:RAIL_snapshots}
\end{figure}

\begin{figure*}
  \centering
  \includegraphics[trim={0cm 4cm 0cm 6cm}, clip, width=0.80\textwidth]{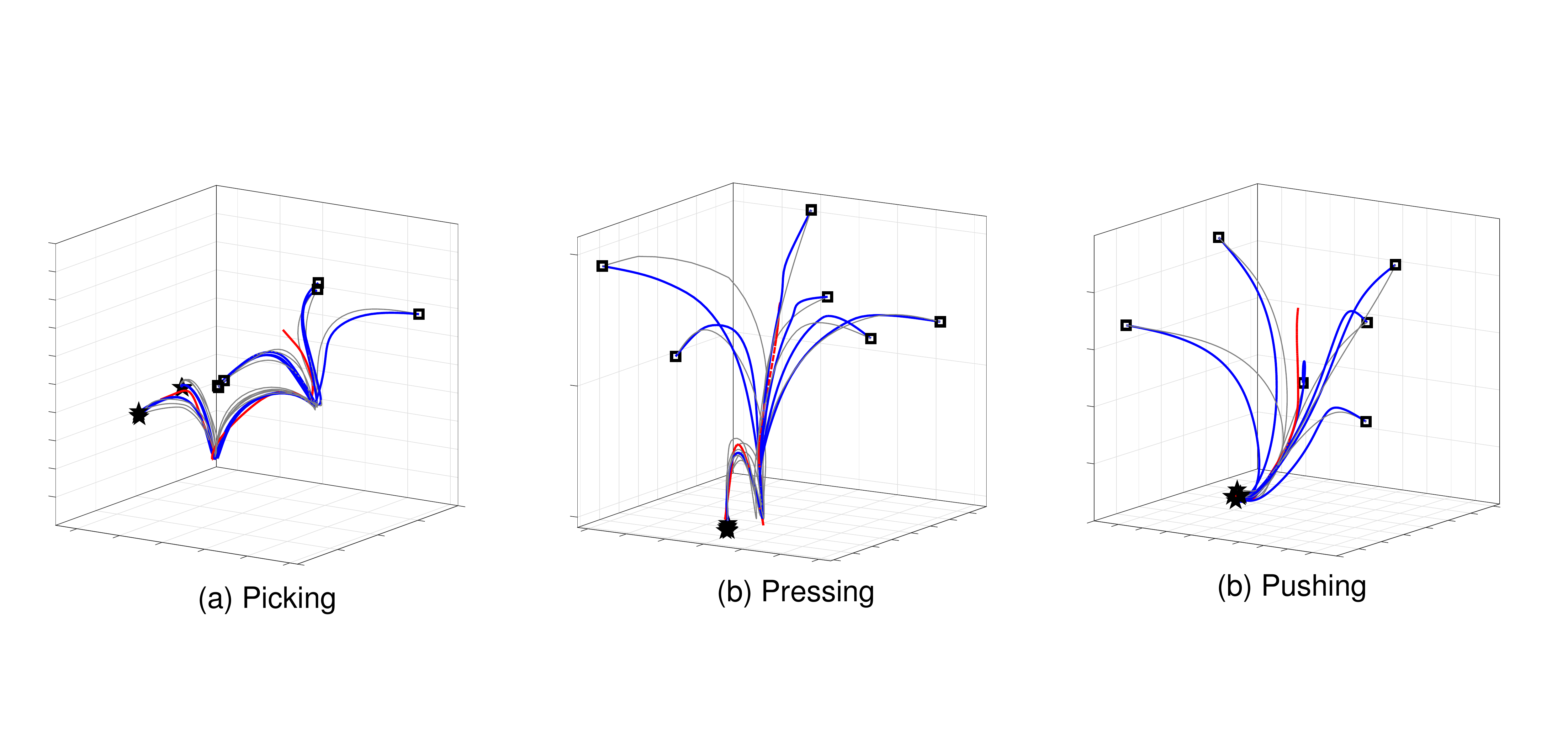}
  \caption{\small{Qualitative performance of MCCB on the picking, pressing, and pushing datasets. Demonstration (gray), reproductions (blue), expected mean position (dashed red), initial (black squares), and target (black stars) are shown.}}
  \label{fig:RAIL_qualitative}
\end{figure*}

Fig. \ref{fig:LASA_qualitative} shows that MCCB yields reproductions that are qualitatively similar to the demonstrations while satisfying the end-point constraints across all motions. As shown in Fig. \ref{fig:LASA_quantitative}, quantitative analysis indicates that MCCB ($\bar{\alpha}_C = 0.1814,\ \bar{\alpha}_G = 0.4958,\ \bar{\alpha}_L = 0.3228$)\footnote{Weighting factors averaged over all 25 skills in the LASA dataset} and three of the four baselines performed comparably with respect to the SEA, FD, SSE, and DTWD metrics, while the Cartesian baseline performed poorly in comparison. This is consistent with the fact that the demonstrations within the LASA dataset emphasize strong similarities in shape.

\subsection{Picking Skill}\label{subsec:RAIL_picking}

The second experiment evaluates the performance of MCCB in a picking task (Fig. \ref{fig:RAIL_snapshots}). The data consists of six kinesthetic demonstrations, each a 3-dimensional robot end-effector position trajectory recorded as a human guided the robot in picking up two magnets atop two blocks. We enforced two via-point constraints (one at each picking point) in addition to the end-point constraints. 

\begin{figure}
  \centering
  \includegraphics[trim={5cm 4cm 4cm 1cm}, clip, width=\columnwidth]{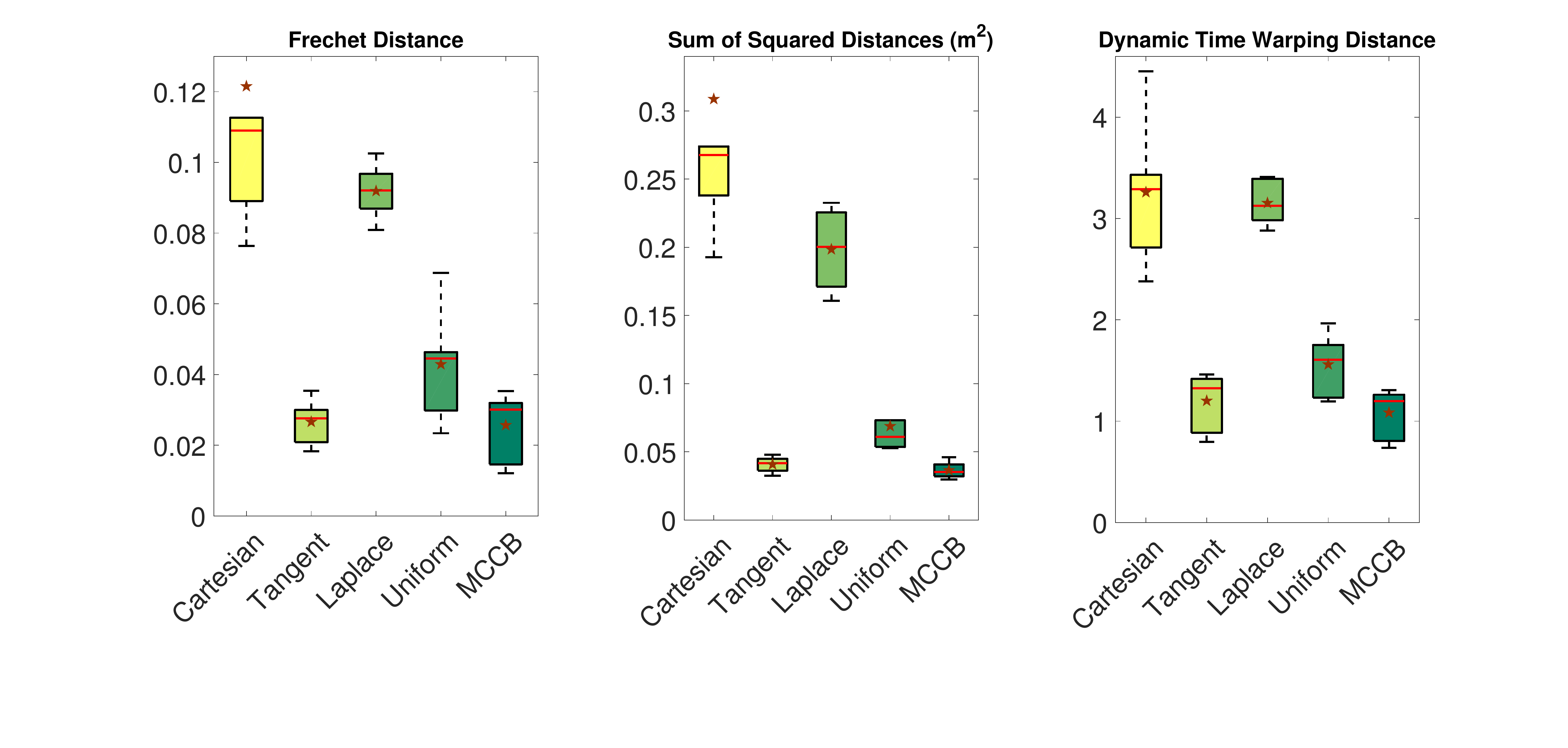}
  \caption{\small{Box plots, with mean (brown star) and median (red line), illustrate the performance of each approach on the picking dataset.}}
  \label{fig:RAIL_picking_quantitative}
\end{figure}

As shown in Fig. \ref{fig:RAIL_qualitative}(a), MCCB generated reproductions that are qualitatively similar to the demonstrations while satisfying all the position constraints. Quantitative evaluations reveal that learning in tangent coordinates yielded better reproductions than learning in Cartesian and Laplacian coordinates (Fig. \ref{fig:RAIL_picking_quantitative}). This was expected since the demonstrations of this task, much like the LASA dataset, emphasize shape similarity. Further, MCCB ($\alpha_C = 0.2362,\ \alpha_G = 0.5451,\ \alpha_L = 0.2187$) yielded the best performance, with respect to all three metrics. In fact, uniform weighting yielded poorer results, with respect to all three metrics, than when considering only the tangent coordinates. The results of this experiment show that while multi-coordinate methods can yield strong performance, it is critical that we balance the weights appropriately.


\subsection{Pressing Skill}\label{subsec:RAIL_pressing}
In this experiment, we evaluated MCCB's ability to learn pressing skills (Fig. \ref{fig:RAIL_snapshots}). The data consists of six kinesthetic demonstrations, each a 3-dimensional robot end-effector position trajectory recorded as a human guided the robot in pressing two cylindrical pegs into their respective holes. 

\begin{figure}
  \centering
  \includegraphics[trim={5cm 4cm 4cm 1cm}, clip, width=\columnwidth]{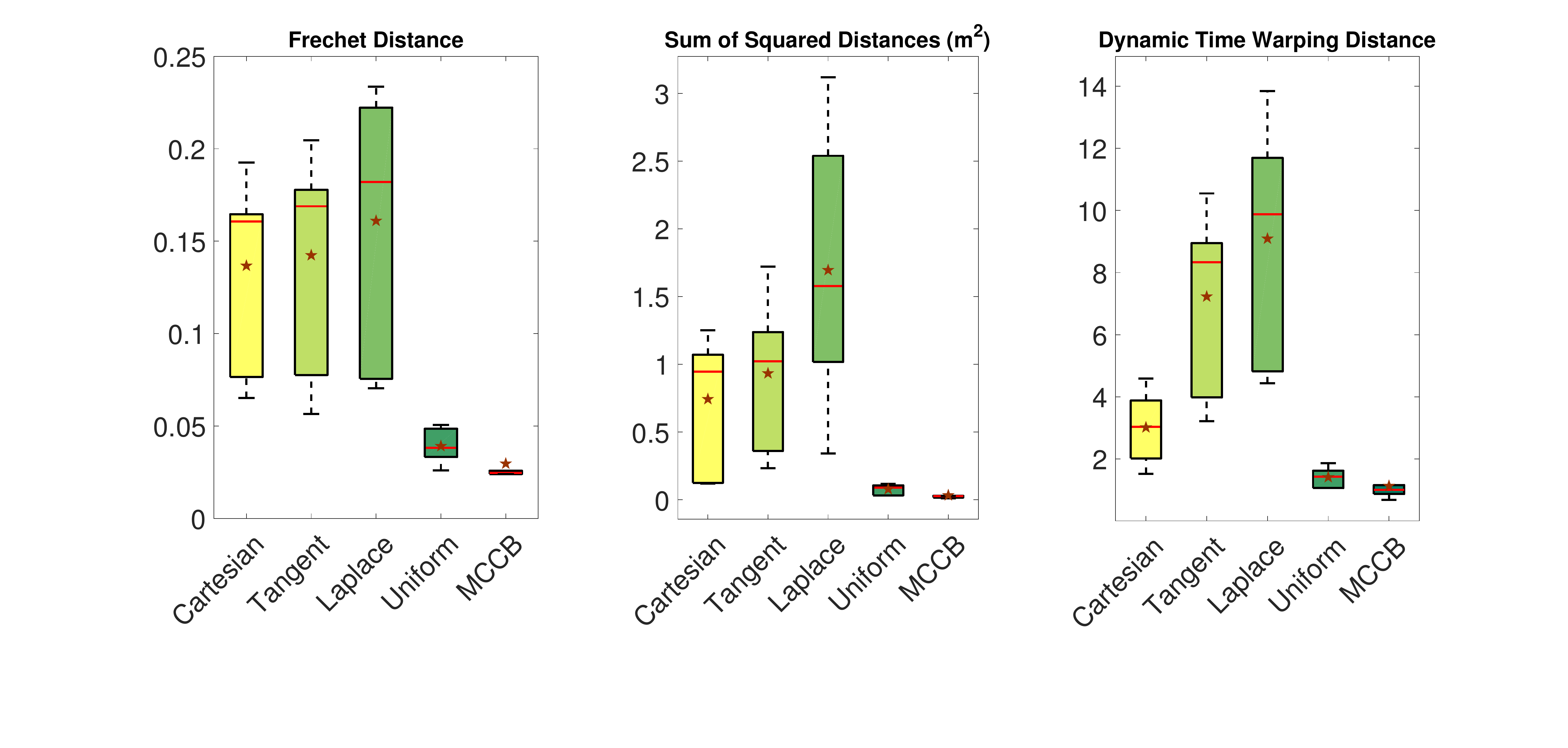}
  \caption{\small{Box plots, with mean (brown star) and median (red line), illustrate the  performance of each approach on the pressing dataset.}}
  \label{fig:RAIL_pressing_quantitative}
\end{figure}

As shown in Fig. \ref{fig:RAIL_qualitative}(b), MCCB successfully reproduced the demonstrations. Note that MCCB is capable of automatically capturing and reproducing the consistencies across the demonstrations in certain regions without any position constraints. Fig. \ref{fig:RAIL_pressing_quantitative} illustrates the performance of MCCB and the baselines with respect to three different metrics. Learning in Cartesian coordinates resulted in the better performance compared to learning in tangent and Laplacian coordinates. Quantitative evaluations further demonstrate that MCCB ($\alpha_C = 0.6735,\ \alpha_G = 0.2034,\ \alpha_L = 0.1231$) consistently yielded the best performance with respect to all three metrics. The results of this experiment, in light of the results in Section \ref{subsec:RAIL_picking}, suggest that the relative importance of each of the differential coordinates vary across different skills.

\subsection{Pushing Skill}\label{subsec 0.045:RAIL_pushing}
\vspace{-.2cm}
The final experiment evaluates the performance of MCCB in a pushing task  (Fig. \ref{fig:RAIL_snapshots}). The data consists of six kinesthetic demonstrations, each a 3-dimensional robot end-effector position trajectory recorded as a human guided the robot in sliding closed the lid of a wooden box. 

\begin{figure}
  \centering
  \includegraphics[trim={5cm 3.9cm 4cm 1cm}, clip, width=\columnwidth]{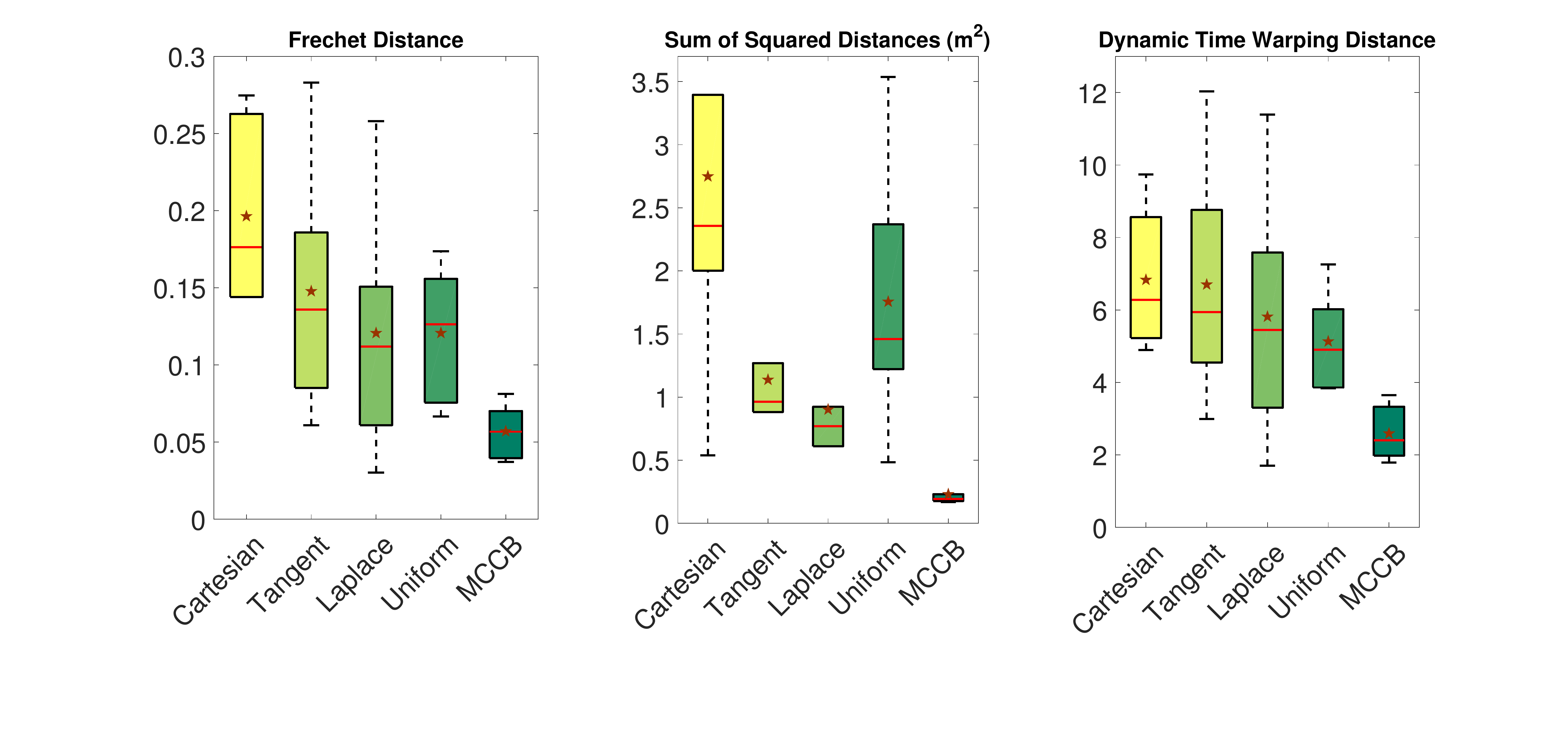}
  \caption{\small{Box plots, with mean (brown star) and median (red line), illustrate the performance of each approach on the pushing dataset.}}
  \label{fig:RAIL_pushing_quantitative}
\end{figure}

As shown in Fig. \ref{fig:RAIL_qualitative}(c), MCCB successfully generated reproductions that are similar to the demonstrations. As evidenced by quantitative evaluations in Fig. \ref{fig:RAIL_pushing_quantitative}, encoding demonstrations in the Laplacian coordinates yielded better performance, with respect to all three metrics, when compared to learning only in either of the other two coordinates, while, MCCB ($\alpha_C = 0.0123,\ \alpha_G = 0.045,\ \alpha_L = 0.9427$) consistently outperformed all the other approaches. Note that learning in the Laplacian coordinates alone resulted in better performance than uniformly weighting of all the coordinates. These results are consistent with the results from the previous sections and indicate that MCCB yields consistently good performance. The results are summarized in Table \ref{tab:summary}.

\begin{table}
\resizebox{\columnwidth}{!}{%
\begin{tabular}{|c"c|c|c"c|c|}
\hline
            & \multicolumn{3}{c"}{Single Coordinate}                                                                                                                 & \multicolumn{2}{c|}{Multi-Coordinate}                                                            \\ \hline
            & Cartesian                                        & Tangent                                          & Laplacian                                        & Uniform W.                                       & MCCB                                       \\ \hline
Handwriting &                                                  & {\color[HTML]{CE6301} \checkmark} {\color[HTML]{009901}  \checkmark} & {\color[HTML]{CE6301} \checkmark} {\color[HTML]{009901} \checkmark} & {\color[HTML]{009901} \checkmark} & {\color[HTML]{009901} \checkmark} \\ \hline
Picking     &                                                  & {\color[HTML]{CE6301} \checkmark} &                                                  &                                                  & {\color[HTML]{009901} \checkmark} \\ \hline
Pressing    & {\color[HTML]{CE6301} \checkmark} &                                                  &                                                  &                                                  & {\color[HTML]{009901} \checkmark} \\ \hline
Pushing     &                                                  &                                                  & {\color[HTML]{CE6301} \checkmark} &                                                  & {\color[HTML]{009901} \checkmark} \\ \hline
\end{tabular}
}
\caption{\small{Orange check marks denote the most relevant coordinate and green check marks denote the best performing method.}}\label{tab:summary}
\end{table}

\section{Discussion and Conclusion}\label{sec:conclusion}
\vspace{-.1cm}

We introduced MCCB, a learning framework for encoding demonstrations in multiple differential coordinates, and automated balancing of costs defined in those coordinates.  As shown in Table \ref{tab:summary}, we demonstrated that the relative effectiveness of each coordinate system is not consistent across a variety of tasks since any given skill might be better suited for learning in one (or more) coordinate system(s). Furthermore, uniform weighting of costs in different coordinates does not consistently yield the best results across different skills. Indeed, uniform weighting, in some cases, yielded poorer performances compared to when only one coordinate system was used. On the other hand, MCCB learned to balance the costs and consistently yielded the best performance. Since the weights are learned directly from the demonstrations, MCCB makes no task-specific assumptions and does not require tedious parameter tuning.  Note that although we used GMMs as the base representation in this work, MCCB is agnostic to the statistical model used to encode the demonstrations in each coordinate system, and thus can be combined with other techniques, such as \cite{calinon2014task,paraschos2013probabilistic,ahmadzadeh2017generalized,umlauft2017learning,rana2017towards,osa2017guiding,nierhoff2016spatial}.  Furthermore, MCCB can be extended to include more coordinate systems that capture additional trajectory features.



\section*{ACKNOWLEDGMENT}

This research is supported in part by NSF NRI 1637758.

\bibliographystyle{IEEEtran}
\bibliography{IEEEabrv,MCCB_ICRA_2019}

\end{document}